\documentclass[journal]{IEEEtran}

\usepackage{cite}
\usepackage{amsmath,amssymb}
\usepackage{newtxtext,newtxmath}
\usepackage{graphicx}
\usepackage{booktabs}
\usepackage{multirow}
\usepackage{array}
\usepackage{placeins}
\usepackage{balance}
\usepackage{url}
\usepackage{xcolor}
\usepackage{microtype}
\usepackage[hidelinks]{hyperref}

\DeclareRobustCommand{\rev}[1]{#1}
\begin{document}

\title{Environment-Invariant Subspace Learning for Generalizable Deepfake Detection}

\author{Shenghao~Chen, Hao~Jia, Chen~Li, Chunjie~Ma,
Zan~Gao,~\IEEEmembership{Senior Member,~IEEE},
and Shengyong~Chen,~\IEEEmembership{Senior Member,~IEEE}%
\thanks{This work was supported in part by the National Natural
Science Foundation of China under Grant 62372325, Grant 62402255,
Grant 62502344, and Grant U25A20444; in part by the Natural Science
Foundation of Tianjin Municipality under Grant 23JCZDJC00280;
in part by the Shandong Provincial Natural Science Foundation under
Grant ZR2024QF020 and Grant ZR2026ZD51; in part by the Shandong
Province National Talents Supporting Program under Grant
2023GJJLJRC070; in part by the Shandong Project towards the
Integration of Education and Industry under Grant 801822020100000024
and Grant 2024ZDZX11; in part by the Young Talent of Lifting
Engineering for Science and Technology in Shandong under Grant
SDAST2024QTB001; and in part by the Open Project Program of the
State Key Laboratory of Virtual Reality Technology and Systems,
Beihang University, under Grant VRLAB2025C05.
(Corresponding author: Zan Gao).}%
\thanks{Shenghao Chen, Hao Jia, Chen Li,  Zan Gao, and
Shengyong Chen are with the Key Laboratory of Computer Vision
and System, Ministry of Education, Tianjin University of Technology,
Tianjin 300384, China.}%
\thanks{Chunjie Ma and Zan Gao are with the Shandong Artificial
Intelligence Institute, Qilu University of Technology
(Shandong Academy of Sciences), Jinan 250014, China.}}

\markboth{Preprint}%
{Chen \MakeLowercase{\textit{et al.}}: Environment-Invariant Subspace Learning for Generalizable Deepfake Detection}

\maketitle

\begin{abstract}
Cross-distribution generalization remains a critical bottleneck in deepfake detection. While recent efforts leverage the semantic priors of large-scale visual foundation models (VFMs), a noteworthy yet underexplored challenge remains: the susceptibility of these semantic priors to environmental interference from factors such as lighting and style. Crucially, this interference establishes spurious correlations between forgery cues and environmental patterns, that severely limit generalization. 
To address this fundamental challenge, we propose an innovative Environment-Invariant Subspace Learning (EISL) framework. The core contribution of EISL is that it aims to disentangle features into orthogonal forgery-relevant invariant factors and environment-related residual factors via a learnable low-rank projection. To facilitate robust feature disentanglement, we also design an Environmental Intervention module that generates diverse and challenging intervention pairs, simulating out-of-distribution environmental shifts to guide the model toward discovering truly invariant forgery representations. \rev{Experiments across cross-dataset, cross-generator, whole-face synthesis, and corruption settings show consistent gains and competitive or leading performance against strong detectors, demonstrating improved robustness to unseen forgery types and environmental variations.} This work provides a new perspective and a valuable exploration for understanding and tackling the generalization barriers of VFMs in deepfake detection.
\end{abstract}


\section{Introduction}
\label{sec:intro}

Following the rapid development of generative artificial intelligence~\cite{difussion_ori,gan_style,diffusion_stable,gan_ori,gan_mmd}, synthetic images are now commonplace. The ease of use of these technologies allows people to create stunningly realistic images, especially facial forgeries~\cite{DeeperForensics,deepfakebench}, that can deceive human observers. Nevertheless, this trend has also raised significant concerns regarding social security. When used for malicious purposes, these technologies can severely undermine public confidence in digital media. Therefore, developing a reliable deepfake detection algorithm is imperative.

One of the central challenges in deepfake detection is enhancing the model's generalization capability to unseen forgery types. Prior work has predominantly relied on paradigms based on specific forgery cues, such as unnatural physiological signals~\cite{biological_signal,head_poses}, spatial domain blending artifacts~\cite{sbi,blending_artifacts1,blending_artifacts2}, and frequency-domain inconsistencies~\cite{spsl,Li_2024_NeurIPS,f3net}. However, detectors tend to overfit to the statistical fingerprints of specific generators present in the training set. Consequently, they struggle with unseen forgeries and exhibit limited generalization to out-of-distribution data. Subsequent studies have attempted to mitigate this by expanding the training distribution through diverse data augmentation~\cite{lsda,face_Augmentation} and disentangled learning \cite{ucf,disentanglement}. Nevertheless, given the rapid iteration of forgery techniques in the wild, these methods remain vulnerable when confronted with images from entirely new generative paradigms.

\begin{figure}[t]
    \centering
    \includegraphics[width=\columnwidth]{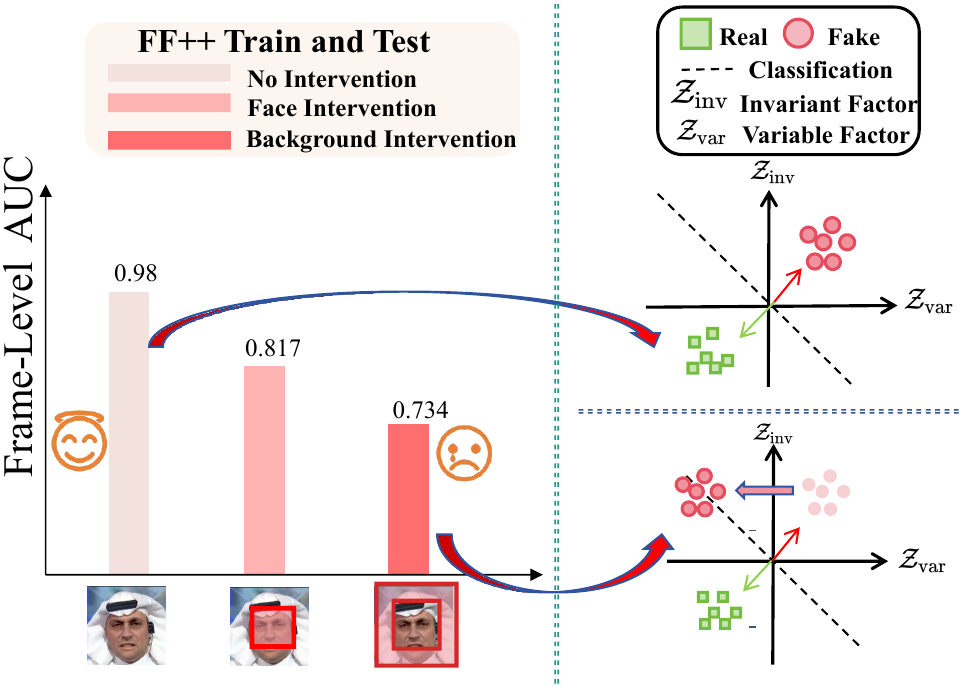}
    \caption{\rev{Motivation. A CLIP ViT-L/14 detector adapted with LoRA on FF++ and evaluated on the FF++ test split retains a frame-level AUC of 0.98 without intervention, but drops to 0.817 and 0.734 when only the facial region or only the background is intervened, although the forgery source and the real/fake label are unchanged.}}
    \label{fig:motivation}
\end{figure}

Recently, the rise of Vision Foundation Models (VFMs) \cite{clip,sam,Caron_2021_ICCV_DINO} has unlocked promising potential for deepfake detection by leveraging the extensive semantic priors of models like CLIP. Conventional approaches typically freeze the CLIP backbone and adopt parameter-efficient fine-tuning (PEFT) methods, such as Adapters~\cite{Houlsby_2019_ICML_Adapters} or LoRA \cite{Hu_2022_ICLR_LoRA}, aiming to exploit these semantic priors to broaden the discriminative boundary against forgeries, and enhancing generalization to unseen manipulation types. However, as noted in \cite{tu2025toward,fang2022data}, CLIP is sensitive to nuisance visual factors including illumination shifts, contrast changes, blur, and compression artifacts, which can substantially degrade its predictions even when high-level semantics remain unchanged. This naturally raises the question: when fine-tuned for deepfake detection, are CLIP's semantic priors susceptible to such uncontrolled environmental variations, thereby undermining cross-domain generalization?


To investigate this, we fine-tuned a CLIP encoder and perturbed the local regions of the FF++ \cite{ff++} dataset (see Fig. \ref{fig:motivation}). The resulting sharp drop in AUC confirms the significant negative impact of environmental interference. Drawing inspiration from CLIP-ICM~\cite{Song_2024_CLIPICM}, we adopt its structural causal perspective to explain the underlying process to partition an image's latent factors into two orthogonal components: Invariant Factors, capturing intrinsic object content or forgery mechanism, and Variant Factors, capturing environmental and contextual variations. When the model's prediction is driven by both factors, it can be easily captured by the variant factors, forcing it to learn spurious correlations. For instance, the model might erroneously associate a specific background or a particular image style with the forgery act itself. Conversely, a prediction mechanism that relies solely on the invariant factors should remain robust across diverse environments. Operations like altering background/foreground colors or image sharpness, for example, do not modify the image's essential content or its ground-truth label (real vs. fake).

To this end, our key insight is that within the learned representation space, we must explicitly disentangle the components susceptible to environmental factors from those genuinely relevant to the forgery detection task. We model the former as an Environmental Residual, while the latter is distilled into an environment-agnostic feature, which is then used for the final forgery discrimination. In this paper, we propose an Environment-Invariant Subspace Learning (EISL) framework to fine-tune CLIP by mitigating semantic interference. Our framework aims to better leverage CLIP's semantic knowledge for robust deepfake generalization. To address environmental variations, we design an Environmental Intervention module that constructs augmented images. Specifically, it generates intervened images by randomly selecting and combining manipulations targeting the foreground, background, shape, and other style attributes. These are then paired with their original counterparts to form environment-intervention pairs. To extract the environment-invariant information, we introduce a learnable low-rank projection matrix that defines an invariant subspace. We then enforce a consistency constraint, ensuring that the projected representations of an intervention pair remain invariant when mapped into this subspace. This process isolates the invariant factors from the environmental residuals, thereby enhancing the model's ability to generalize by focusing only on the fundamental attributes of the forgery. Our main contributions can be summarized as follows:

\begin{enumerate}
    
    \item We reveal that semantic priors in  Vision Foundation Models (VFMs) for deepfake detection are vulnerable to environmental factors such as background, style, and illumination, which easily distract detectors and lead to poor cross-domain generalization.
    
    \item We propose an Environment-Invariant Subspace Learning (EISL) framework to address this problem. By employing a learnable low-rank projection to decouple forgery-relevant invariant components from environmental residuals, EISL effectively adapts VFMs for robust and environment-insensitive deepfake detection.
    
    \item \rev{Extensive experiments across cross-dataset, cross-generator, whole-face synthesis, and corruption settings show consistent improvements and competitive or leading performance, supported by mechanism analyses of projected and residual feature shifts.}
\end{enumerate}

\section{Related Work}
\label{sec:related}

\subsection{Deepfake Detection based on Cues.}
Deepfake Detection is a central problem in digital manipulation forensics. With the rise of deep learning and the strong local modeling capacity of CNNs~\cite{efficientnet,x-ray,ff++}, many methods focus on explicit forgery cues for classification. Typical cues include physiological and temporal inconsistencies~\cite{biological_signal,head_poses,tifs_istvt} (e.g., blinking patterns, rPPG, audio-visual or lip-sync mismatch), frequency-domain artifacts~\cite{spsl,Li_2024_NeurIPS,f3net,tifs_sficonv} (e.g., compression traces, resampling, phase anomalies), geometric and boundary distortions~\cite{sbi,blending_artifacts1,blending_artifacts2,blending_artifacts3} (e.g., abnormal pose, landmark distribution, blending seams), as well as semantic and noise-level telltales~\cite{tifs_detect_locate}. Although these approaches often achieve excellent in-domain performance, they heavily rely on carefully curated training data and tend to overfit to seen forgery patterns, thus struggling to generalize to unknown forgeries~\cite{tifs_lisiam,tifs_masked_relation,tifs_fedforgery,tifs_gradient_reg}. Subsequent works attempt to alleviate this by synthesizing pseudo-forgeries to expand the decision boundary. For example, LSDA~\cite{lsda} constructs intra- and inter-forgery variations in the latent space to enlarge the forgery space; FIA-USA~\cite{ma2025from} randomly selects facial sub-regions and reconstructs them to simulate upsampling inconsistencies; FreqDebias~\cite{Kashiani_2025_CVPR} injects perturbations into gradient-sensitive frequency bands to enhance sensitivity to unseen artifacts. However, due to the limited diversity of training forgeries and the capacity of the adopted backbones, it remains difficult for these methods to learn truly universal forgery-discriminative features, and a clear performance gap persists under unknown manipulations.

\subsection{Deepfake Detection based on Knowledge.}
Recently, with the success of large vision-language models, an emerging line of work exploits CLIP-like semantic knowledge to improve cross-domain generalization in deepfake detection. FFAA~\cite{ffaa} constructs a dataset with textual reasoning to support multi-answer decision making; VLFFD~\cite{vlffd} generates prompt-guided synthetic images for joint training; RepDFD~\cite{repdfd} proposes a reprogramming strategy that injects universal perturbations into visual inputs while keeping VLM parameters frozen. CLIPping~\cite{clipping} further benchmarks CLIP with various adapter modules for universal detection. ForAda~\cite{cui2025forensics} fine-tunes CLIP via adapters with combined contrastive and boundary losses to enhance generalization, while Effort~\cite{yan2025effort} applies SVD-based low-rank adaptation on CLIP weights to achieve strong performance with very few trainable parameters. However, existing methods do not consider that semantic knowledge can struggle under environmental interference, leading to suboptimal performance. Our approach seeks to identify causally invariant, forgery-discriminative cues in images, thereby enabling generalized detection when leveraging pre-trained knowledge.

\section{Method}
\label{sec:method}

\subsection{Overview}
\begin{figure*}[t]
    \centering
    \includegraphics[width=\textwidth]{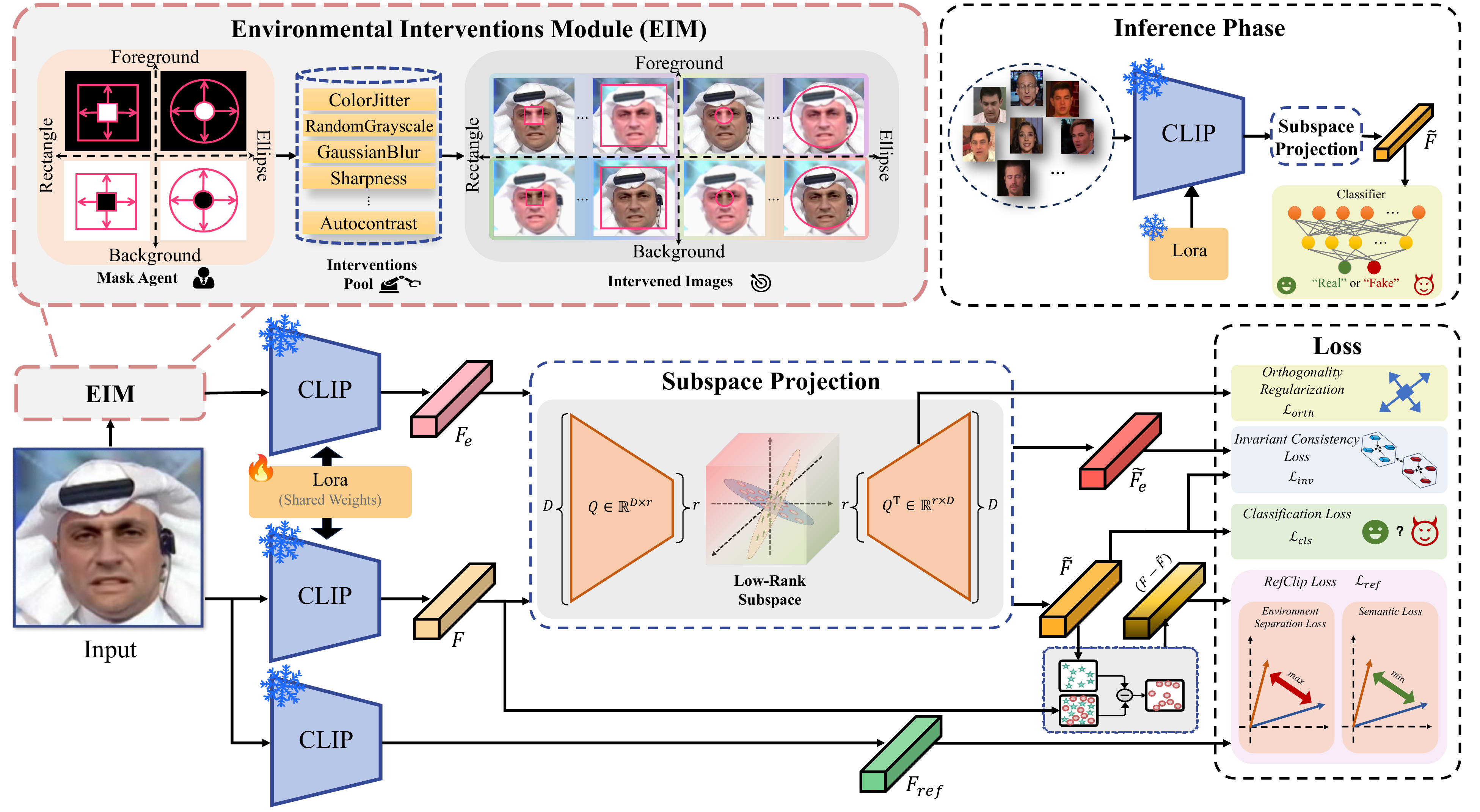}
    \caption{EISL framework. During training, EIM generates a label-preserving environmental intervention $I_e$. Original and intervened images are encoded by a shared LoRA-adapted CLIP, and a frozen CLIP provides the semantic reference. The low-rank projector produces invariant features for classification and consistency learning; the complementary residual is separated from the reference semantics. At inference, only the adapted CLIP, projector, and classifier are used.}
    \label{fig:framework}
\end{figure*}

Recent works~\cite{cui2025forensics,yan2025effort} leverage CLIP-based encoders for deepfake detection to benefit from strong semantic priors and cross-domain generalization. However, CLIP features are highly sensitive to environmental factors such as illumination, style, and compression~\cite{tu2025toward,fang2022data}. Fine-tuning on specific datasets amplifies this sensitivity, causing detectors to overfit dataset environments rather than intrinsic forgery cues, leading to severe degradation under cross-dataset evaluation.
We address this with Environment-Invariant Subspace Learning (EISL), which constrains decisions to a compact, environment-robust subspace of CLIP features. The overall framework is illustrated in Fig.\ref{fig:framework}. Specifically, we construct Environmental Interventions generating label-preserving pairs with realistic environment shifts, and learn a low-rank projection mapping CLIP features to an environment-invariant subspace. The projection is optimized with four objectives: semantic alignment (preserving forgery semantics), residual separation (preventing environmental leakage), interventional consistency (enforcing robustness to shifts), and orthogonality regularization (stabilizing the basis).

\subsection{\rev{Operational Definition of Environmental Factors}}
\label{sec:env_def}
\rev{In this work, an \emph{environmental factor} is a label-preserving, non-forgery variation caused by acquisition, post-processing, or visual context. Operationally, it may alter illumination, color, contrast, blur, sharpness, compression-like statistics, or foreground/background appearance, but it must preserve the depicted identity, the forgery source, and the binary real/fake label. Identity or content edits and changes to the manipulation mechanism are therefore outside this definition. This criterion makes the nuisance family testable: an original image and its intervention share the forensic target, while their feature difference exposes sensitivity that the detector should not use.}

\subsection{Environmental Interventions Module (EIM)}
\label{subsec:envinterv}
\label{sec:eim}

A natural choice for environmental interventions in deepfake detection is to rely on accurate facial masks. One may perturb the facial region while keeping the background fixed, or conversely modify the background while preserving the face. However, such deterministic localization inevitably encodes a strong spatial prior: perturbations and potential manipulations are consistently tied to a fixed facial region with well-defined boundaries. When used for training, this prior introduces trivial location cues (e.g., mask-shaped transitions) that models can exploit, rather than encouraging representations that remain reliable under diverse manipulation patterns and acquisition conditions. Our objective is therefore to construct interventions that expose the model to diverse and realistic environmental changes while avoiding shortcut cues induced by fixed masks or deterministic boundaries. Given an input image $I \in \mathbb{R}^{3 \times H \times W}$ (real or fake), our Environmental Interventions Module (EIM) produces an intervened image $\tilde{I}$ such that the identity and real/fake label remain unchanged, whereas environmental factors are explicitly and controllably perturbed. EIM encompasses the following steps:

\noindent\textbf{Random Center Mask Generation.  } 
Given an input image $I \in \mathbb{R}^{3 \times H \times W}$, we first sample a shape $s \in \{\mathrm{rect}, \mathrm{ellipse}\}$,
then we independently sample horizontal and vertical scale factors $k_h, k_w \sim U(k_{\min}, k_{\max})$,
where in our experiments $k_{\min}=0.2$ and $k_{\max}=0.8$. The sampled scales determine the spatial extent of the central mask, specifically, we set $h = \lfloor H k_h \rfloor, w = \lfloor W k_w \rfloor$, where $h$ and $w$ denote the mask height and width, respectively. Let $(c_x, c_y)$ denote the image center. We construct a binary mask $M \in \{0,1\}^{H \times W}$ by assigning, for each pixel location $(x,y):$
\begin{equation}
M(y,x) =
\begin{cases}
1, & s = \mathrm{rect},\ |y-c_y|\le \tfrac{h}{2},\ |x-c_x|\le \tfrac{w}{2}, \\[4pt]
1, & s = \mathrm{ellipse},\
\bigl(\tfrac{x-c_x}{w/2}\bigr)^2 + \bigl(\tfrac{y-c_y}{h/2}\bigr)^2 \le 1, \\[4pt]
0, & \text{otherwise},
\end{cases}
\label{eq:mask}
\end{equation}
pixels with $M(y,x)=1$ define the central region of interest, while pixels with
$M(y,x)=0$ form the complementary context region. Since $h$, $w$, and $s$ are sampled
independently for each image, the intervened region only probabilistically overlaps
the facial area and does not induce a deterministic manipulation boundary. This
stochastic design is crucial for avoiding spatial shortcut cues in the subsequent
environmental intervention.

\noindent\textbf{Interventional Blending. }
To introduce environment-level variability without altering the underlying identity or forgery label, we combine strong appearance augmentation with
stochastic region-wise blending guided by the central mask $M$. We employ transformations that primarily affect low-level appearance statistics (e.g., color tone, contrast, blur) while preserving the spatial layout and forgery cues, so that the forgery patterns remain intact and the binary real/forged label is unchanged. In this way, pairs $(I, I')$ constructed by EIM are label-consistent yet environmentally diverse, encouraging the detector to produce stable predictions
under contextual changes and to treat such environmental variations as nuisance
factors rather than discriminative shortcuts. Specifically, we first define a set of candidate Interventions Pool $\mathcal{A}$ = \{\text{ColorJitter}, \text{RandomGrayscale}, \text{GaussianBlur}, \text{Sharpness}, \text{Autocontrast}\}. For each image $I$, we sample a non-empty random subset of $\mathcal{A}$ and compose them into a transformation $T$, yielding an appearance-shifted variant: 
\begin{equation}
I_{\text{aug}} = T(I),
\label{eq:aug}
\end{equation}
where $I_{\text{aug}}$ is the augmented image, these transformations primarily modify illumination, color characteristics, and
low-level textures, mimicking variations in capture devices and conditions while
preserving high-level semantics. Given the binary mask $M$ from Eq.~\eqref{eq:mask}, we then construct the intervened image $I'$ by randomly perturbing either the central region or its
complementary context.
We draw $b \sim \text{Bernoulli}(0.5)$ and define:
\begin{equation}
I' =
\begin{cases}
M \odot I + (1 - M) \odot I_{\text{aug}}, & \text{if } b = 1 ,\\
M \odot I_{\text{aug}} + (1 - M) \odot I, & \text{if } b = 0 ,
\end{cases}
\label{eq:blend}
\end{equation}
where $\odot$ denotes element-wise multiplication.
Thus, either the central region or the surrounding region is replaced by its
strongly augmented counterpart, but never both simultaneously. This interventional blending exposes the model to diverse combinations of foreground and background appearance, while the stochastic mask design prevents a fixed association between specific spatial locations and environmental
perturbations. As a result, EIM provides rich, label-preserving environment shifts
without introducing deterministic spatial shortcuts.

\subsection{Environment-Invariant Subspace Construction}
\label{sec:eisl_subspace}
\label{sec:subspace}

\paragraph{Motivation and Design Principles.} Given a face image $I$ and its environment-perturbed counterpart $I'$, we extract their features using a LoRA-tuned CLIP encoder $\Phi_{\theta}(\cdot)$:
\begin{equation}
F=\Phi_{\theta}(I),\quad F_{e}=\Phi_{\theta}(I') \in \mathbb{R}^{D},
\label{eq:features}
\end{equation}
where these features inevitably entangle forgery cues with environmental factors. A naive approach would apply a fully-connected layer $Z=W\!F$ for classification. However, such unconstrained linear transformation will freely mix all dimensions and amplify spurious, environment-specific signals, degrading cross-distribution generalization.

To suppress environment-sensitive directions while preserving transferable forgery cues, we propose a low-rank subspace projector that geometrically separates environment-invariant forgery representations from nuisance environmental variations. Specifically, our subspace projector design is guided by three key principles.
Firstly, unlike arbitrary linear transformation, our projector is symmetric and approximately idempotent, acting as a geometric projector onto a subspace rather than an arbitrary feature mixer. This projection geometry mechanism naturally suppresses out-of-subspace environmental noise.
Secondly, we impose a low-rank bottleneck that creates an information compression effect. By restricting the subspace dimension to be much smaller than the feature dimension, we force the model to prioritize robust, transferable forgery patterns while discarding environment-specific variations.
Thirdly, we advocate a shared task-aligned subspace, in which the same projection is applied to both original and perturbed inputs, and jointly optimized for invariance and semantics objectives, explicitly encouraging the subspace to capture environment-agnostic encoding.

\paragraph{Subspace Projection Formulation.} 
We realize these principles through a learnable low-rank matrix 
$Q\in\mathbb{R}^{D\times r},\quad r\ll D$,
regularized to be column-orthonormal (Sec. \ref{sec:objective}). This defines our projection operator:
\begin{equation}
P=QQ^{\top}\in\mathbb{R}^{D\times D}.
\label{eq:projector}
\end{equation}
The orthonormality of $Q$ ensures $P$ satisfies the projection properties: $P=P^{\top}$ (symmetry) and $P^{2}\!\approx\!P$ (idempotence), while the low-rank constraint $\mathrm{rank}(P)\le r$ enforces the information bottleneck. This projector transforms features into an environment-invariant subspace:
\begin{equation}
\widetilde{F}=PF,\quad \widetilde{F}_{e}=PF_{e},
\label{eq:projection}
\end{equation}
where applying the same projection to both $I$ and $I'$ enforces that downstream decisions lie in a shared, compact subspace, attenuating nuisance variation and retaining stable, forgery-relevant structure.
All classification operates on $\tilde{F}$, steering the detector toward stable manipulation cues instead of raw, environment-entangled features. 

Compared to conventional fully-connected layers, our projector enforces explicit geometric properties, restricts capacity through low-rank factorization, and uses a shared projection jointly optimized for invariance; collectively, these constraints enable robust environment-agnostic feature learning.
{\color{black}
\paragraph{Geometric and Invariance Guarantees.}
\noindent\textit{Proposition 1 (orthogonal projection).}
If $Q^\top Q=I_r$, then $P=QQ^\top$ is the orthogonal projector onto $\mathcal{S}=\mathrm{span}(Q)$, and every feature has the canonical decomposition
\begin{equation}
F=PF+(I_D-P)F,\qquad P(I_D-P)=0.
\label{eq:orth_decomp}
\end{equation}
Indeed, $P^\top=P$ and $P^2=Q(Q^\top Q)Q^\top=P$. Hence $P$ preserves vectors in $\mathcal{S}$ and annihilates vectors in $\mathcal{S}^{\perp}$; the projected feature and residual are therefore geometrically complementary rather than an arbitrary arithmetic split. During training, $\mathcal{L}_{\mathrm{orth}}$ makes this property approximate to the extent that $Q^\top Q$ approaches $I_r$.

\noindent\textit{Proposition 2 (rank bottleneck).}
The projected feature lies in a subspace of dimension at most $r$, because
\begin{equation}
\mathrm{rank}(P)=\mathrm{rank}(QQ^\top)\leq \mathrm{rank}(Q)\leq r.
\label{eq:rank_bound}
\end{equation}
This constraint limits the capacity available for environment-dependent shortcuts. Although EISL shares LoRA's parameter-efficiency motivation~\cite{Hu_2022_ICLR_LoRA}, their operators have different roles: a LoRA update $BA$ changes a network weight and is generally asymmetric and non-idempotent, whereas $QQ^\top$ is a symmetric feature-space filter whose input and output subspaces coincide.

\noindent\textit{Proposition 3 (invariance from paired interventions).}
For analysis, write a label-preserving pair locally as $F=s+e$ and $F_e=s+e_e$, where $s$ is shared and $e,e_e$ are environment-dependent components. Then
\begin{equation}
\mathcal{L}_{\mathrm{inv}}
=\|P(F-F_e)\|_2^2
=\|P(e-e_e)\|_2^2.
\label{eq:invariance_mechanism}
\end{equation}
Thus, zero consistency loss for an observed difference $e-e_e$ implies $e-e_e\in\ker(P)=\mathcal{S}^{\perp}$. Over multiple pairs, minimizing the empirical objective suppresses the span of observed intervention differences inside $\mathcal{S}$ and routes it toward the complementary residual. This is a conditional mechanism rather than a claim that every open-world nuisance is removed: its coverage depends on the interventions, finite data, approximate orthogonality, and the competing requirements of classification and semantic preservation.
}

\subsection{Objective Function}
\label{sec:objective}

Our overall objective is to train a detector capable of disentangling forgery-relevant invariant semantics from environment-specific factors, while simultaneously preserving the discriminative cues essential for deepfake detection. To achieve this, we formulate an overall optimization objective $\mathcal{L}_{\text{total}}$, which is composed of a primary binary classification loss and a weighted sum of three complementary constraint losses.

\noindent\textbf{Binary Classification Loss. }
To finally achieve the deepfake detection task, we add a binary classifier to the encoder (based on its invariant feature $\tilde{F}$) to distinguish between real and fake samples. This classification loss, the standard Binary Cross-Entropy (BCE) Loss, is formulated for a single sample as:
\begin{equation}
\mathcal{L}_{\text{cls}} = - [y \log(p) + (1-y) \log(1-p)],
\label{eq:cls}
\end{equation}
where $y \in \{0, 1\}$ is the ground-truth label (with 1 denoting fake), and $p = \text{MLP}(\tilde{F})$ is the model's predicted probability that the sample is fake.

\noindent\textbf{RefCLIP Constraint Loss. }
During the projection from the CLIP feature space to the environment-invariant subspace,  we additionally use a frozen CLIP encoder to extract a reference feature $F_{ref}$, this loss acts as a reference-model constraint: it explicitly keeps the invariant feature $\tilde{F}$ close to the facial semantic space defined by the reference CLIP feature $F_{\text{ref}}$, while enforcing the environmental residual $F - \tilde{F}$ to be orthogonal to this semantic information.


To achieve this, we first define the Semantic Alignment Loss $\mathcal{L}_{\text{align}}$, which uses cosine similarity to pull $\tilde{F}$ and $F_{\text{ref}}$ closer:
\begin{equation}
\mathcal{L}_{\text{align}} = 1 - \mathrm{sim}(\tilde{F}, F_{\text{ref}}),
\label{eq:align}
\end{equation}
second, we define the Environment Separation Loss $\mathcal{L}_{\text{sep}}$ to penalize correlation between the environment residual $F - \tilde{F}$ and $F_{\text{ref}}$:
\begin{equation}
\mathcal{L}_{\text{sep}} =  |\mathrm{sim}(F - \tilde{F}, F_{\text{ref}})|,
\label{eq:orth}
\end{equation}
where $\mathrm{sim}(\cdot,\cdot)$ denotes cosine similarity. Finally, the RefCLIP constraint is defined as the combination of these two terms:
\begin{equation}
\mathcal{L}_{\text{ref}} = \mathcal{L}_{\text{align}} + \mathcal{L}_{\text{sep}},
\label{eq:refclip}
\end{equation}

\noindent\textbf{Invariant Consistency Loss. }
 This loss is key to improving the model's generalization ability. We generate a label-preserving environmental view of each image and require its invariant feature $\widetilde{F}_{e}$ to remain consistent with the original invariant feature $\widetilde{F}$. This alignment is quantified using squared L2 distance:
\begin{equation}
\mathcal{L}_{\text{inv}} = \left\| \widetilde{F} - \widetilde{F}_{e} \right\|_{2}^{2},
\label{eq:inv}
\end{equation}
This loss forces the model to learn representations that are stable under various environmental perturbations.

\noindent\textbf{Low-Rank Orthogonality Regularization. }
 To ensure that the low-rank basis $Q \in \mathbb{R}^{D \times r}$ used for projection forms a well-structured, non-degenerate subspace, we impose a regularization term. This term penalizes the deviation of $Q$'s column vectors from an orthonormal basis:
\begin{equation}
\mathcal{L}_{\text{orth}} = \left\| Q^\top Q - I_r \right\|_{F}^{2},
\label{eq:sub}
\end{equation}
where $I_r$ is the $r$-dimensional identity matrix and $\|\cdot\|_F$ is the Frobenius norm. This helps to stabilize the disentanglement process.

\noindent\textbf{Overall Loss. }
 The final loss function is obtained by the weighted sum of all the above loss terms:
\begin{equation}
\mathcal{L}_{\text{total}}
= \mathcal{L}_{\text{cls}}
+ \lambda_{\text{ref}} \mathcal{L}_{\text{ref}}
+ \lambda_{\text{inv}} \mathcal{L}_{\text{inv}}
+ \lambda_{\text{orth}} \mathcal{L}_{\text{orth}}.
\label{eq:total}
\end{equation}
where $\lambda_{\text{ref}}$, $\lambda_{\text{inv}}$, and $\lambda_{\text{orth}}$ are hyper-parameters balancing the contribution of each term. In our experiments, we empirically set $\lambda_{\text{ref}}=0.1$, $\lambda_{\text{inv}}=0.2$, and $\lambda_{\text{orth}}=0.01$.

\paragraph{\rev{Role and Interaction of the Objective Terms.}}
\label{sec:role}
\rev{The four terms prevent different failure modes and should not be interpreted as interchangeable. $\mathcal{L}_{\mathrm{cls}}$ preserves label-discriminative directions and prevents a trivial collapsed representation; $\mathcal{L}_{\mathrm{inv}}$ identifies the pairwise-unstable directions that should leave the projected subspace; and $\mathcal{L}_{\mathrm{orth}}$ supplies the projector geometry required by Propositions~1 and~3. RefCLIP is only a semantic anchor: its cosine constraints preserve the direction of useful facial semantics and discourage their leakage into the residual, but they do not themselves remove environmental variation. The robustness mechanism is therefore the coupling of paired consistency with the low-rank projector, while RefCLIP limits semantic loss. The component study in Table~\ref{tab:ablation} empirically separates these roles.}

\section{Experiments}
\label{sec:experiments}

\subsection{Experimental Setup}
\noindent\textbf{Datasets and protocol:} We evaluate our approach on eight widely used deepfake datasets: FaceForensics++ (FF++)~\cite{ff++}, the Deepfake Detection Challenge (DFDC)~\cite{dfdc}, the DFDC preview set (DFDCP)\cite{dfdcp},  two versions of CelebDF (CDF-v1, CDF-v2)\cite{celeb},  DeepFakeDetection dataset (DFD)\cite{dfd}, DF40~\cite{yan2024df40} and Diffusion Facial Forgery (DiFF)~\cite{cheng2024diff}. 

\noindent\textbf{Implementation:} Our method is based on PyTorch~2.1.2 and runs on an NVIDIA Tesla A100 GPU.
We adopt the CLIP-Large-Patch14 vision encoder, initialize it with pretrained weights, and keep it frozen
while learning forgery-discriminative representations through LoRA-based adaptation ($r_{\text{lora}}=16$).
Input face images are resized to \(224 \times 224\), and the model is trained for 10 epochs using the Adam optimizer
with a learning rate of \(1\times10^{-5}\) and a batch size of 16.
The rank for the low-rank subspace is set to \(r = 64\).
Following the configuration of DeepfakeBench \cite{deepfakebench}, we train on the c23-compressed version of FF++.
For each video, 8 frames are sampled for training and 32 frames for testing.

\subsection{Cross-Dataset and Cross-Generator Generalization}
\begin{table*}[!t]
\caption{Cross-dataset frame-level AUC. All methods are trained on FF++ and evaluated on unseen datasets. Best and second-best values are bold and underlined, respectively.}
\label{tab:frame}
\centering
\setlength{\tabcolsep}{5.2pt}
\begin{tabular}{lccccccc}
\toprule
Method & Venue & CDF-v1 & CDF-v2 & DFDC & DFDCP & DFD & Avg.\\
\midrule
Xception~\cite{ff++} & ICCV'19 & 0.779 & 0.737 & 0.708 & 0.737 & 0.816 & 0.755\\
EfficientB4~\cite{efficientnet} & ICML'19 & 0.791 & 0.749 & 0.696 & 0.728 & 0.815 & 0.756\\
F3Net~\cite{f3net} & AAAI'20 & 0.777 & 0.735 & 0.702 & 0.735 & 0.798 & 0.749\\
X-ray~\cite{x-ray} & CVPR'20 & 0.709 & 0.679 & 0.633 & 0.694 & 0.766 & 0.696\\
FFD~\cite{ffd} & CVPR'20 & 0.784 & 0.744 & 0.703 & 0.743 & 0.802 & 0.755\\
SPSL~\cite{spsl} & CVPR'21 & 0.815 & 0.765 & 0.704 & 0.741 & 0.812 & 0.767\\
SRM~\cite{srm} & CVPR'21 & 0.793 & 0.755 & 0.700 & 0.741 & 0.812 & 0.760\\
Recce~\cite{recce} & CVPR'22 & 0.768 & 0.732 & 0.713 & 0.734 & 0.812 & 0.752\\
SBI~\cite{sbi} & CVPR'22 & -- & 0.813 & -- & 0.799 & 0.774 & --\\
UCF~\cite{ucf} & ICCV'23 & 0.779 & 0.753 & 0.719 & 0.759 & 0.807 & 0.763\\
ED~\cite{ba} & AAAI'24 & 0.818 & 0.864 & 0.721 & 0.851 & -- & --\\
LSDA~\cite{lsda} & CVPR'24 & 0.867 & 0.830 & 0.736 & 0.815 & 0.880 & 0.826\\
CFM~\cite{cfm} & TIFS'24 & -- & 0.828 & -- & 0.758 & 0.915 & --\\
ForAda~\cite{cui2025forensics} & CVPR'25 & \underline{0.914} & \underline{0.900} & \textbf{0.843} & \underline{0.890} & \underline{0.933} & \underline{0.896}\\
FIA-USA~\cite{ma2025from} & NeurIPS'25 & 0.901 & 0.867 & -- & 0.818 & 0.821 & 0.852\\
\midrule
EISL (ours) & -- & \textbf{0.920} & \textbf{0.911} & \underline{0.833} & \textbf{0.894} & \textbf{0.943} & \textbf{0.900}\\
\bottomrule
\end{tabular}
\end{table*}


For a fair comparison, we evaluate our method and all competing end-to-end approaches at both frame-level and video-level. We adopt the unified generalization evaluation protocol of DeepfakeBench~\cite{deepfakebench} to ensure consistent data preprocessing, experimental settings, and metrics across datasets. For recent state-of-the-art methods that are already evaluated under the DeepfakeBench configuration, we directly use their reported results for comparison.

\smallskip
\noindent\textbf{Frame-level Comparison. } 
We treat each extracted face frame as an individual sample and aggregate predictions over all frames in the test videos to compute image-level detection metrics. Table \ref{tab:frame} reports the cross-dataset performance in terms of frame-level AUC, which reflects the ability of different methods to distinguish real from fake content using static visual evidence. Our method achieves the best performance on CDF-V1, CDF-V2, DFDCP, and DFD, and ranks second only to ForAda on DFDC, demonstrating strong generalization when trained with subspace learning. To further evaluate the generalization of forgery types, following~\cite{ma2025from}, we select uniface, e4s, FaceDancer, FSGAN, InSwap, and SimSwap as representative cross-domain manipulation methods on the DF40 dataset, as shown in Table  \ref{tab:df40}, our approach is consistently ranked first or second across all these manipulation types and achieves an average improvement of 1.3\% over all data, further confirming the robustness and transferability of the proposed framework to diverse and unseen manipulation patterns.

%

\smallskip
\noindent\textbf{Video-level Comparison. }
We obtain predictions by averaging detection scores over all sampled frames of each video.
As reported in Table \ref{tab:video}, our method achieves the best video-level AUC on CDF-v2 (0.969) and DFDCP (0.940), outperforming both previous video-based architectures and recent frame-based detectors.
On DFDC, our approach also remains highly competitive with an AUC of 0.856, surpassing most existing methods and only slightly below ForAda, the results strongly demonstrate the effective generalization of our method.

\begin{table}[!tb]
\caption{Cross-dataset video-level AUC. The upper block contains video-based methods and the lower block frame-based methods.}
\label{tab:video}
\centering
\setlength{\tabcolsep}{4.2pt}
\begin{tabular}{lcccc}
\toprule
Method & Venue & CDF-v2 & DFDC & DFDCP\\
\midrule
RealForensics~\cite{RealForensics} & CVPR'22 & 0.869 & 0.759 & --\\
TALL~\cite{tall} & ICCV'23 & 0.908 & 0.768 & --\\
AltFreezing~\cite{altfreezing} & CVPR'23 & 0.895 & -- & --\\
SeeABLE~\cite{SeeABLE} & ICCV'23 & 0.873 & 0.759 & 0.863\\
IID~\cite{iid} & CVPR'23 & 0.838 & -- & 0.812\\
TALL++~\cite{tall++} & IJCV'24 & 0.920 & 0.785 & --\\
SAM~\cite{sam} & CVPR'24 & 0.890 & -- & --\\
VB~\cite{YanZCGFYDWY25} & CVPR'25 & 0.947 & 0.843 & 0.909\\
\midrule
SBI~\cite{sbi} & CVPR'22 & 0.932 & 0.724 & 0.862\\
AUNet~\cite{aunet} & CVPR'23 & 0.928 & 0.738 & 0.862\\
CADDM~\cite{caddm} & CVPR'23 & 0.939 & 0.739 & --\\
SFDG~\cite{sfgd} & CVPR'23 & 0.758 & 0.736 & --\\
LAA-NET~\cite{laa-net} & CVPR'24 & 0.954 & -- & 0.869\\
LSDA~\cite{lsda} & CVPR'24 & 0.911 & 0.770 & --\\
CFM~\cite{cfm} & TIFS'24 & 0.897 & -- & 0.802\\
ForAda~\cite{cui2025forensics} & CVPR'25 & \underline{0.957} & \textbf{0.872} & \underline{0.929}\\
Effort~\cite{yan2025effort} & ICML'25 & 0.956 & 0.843 & 0.909\\
FIA-USA~\cite{ma2025from} & NeurIPS'25 & 0.941 & 0.732 & 0.866\\
\midrule
EISL (ours) & -- & \textbf{0.969} & \underline{0.856} & \textbf{0.940}\\
\bottomrule
\end{tabular}
\end{table}

\begin{table*}[!t]
\color{black}
\caption{Frame-level AUC (\%) on six representative DF40 face-swapping methods. All models are trained on FF++.}
\label{tab:df40}
\centering
\setlength{\tabcolsep}{5pt}
\begin{tabular}{lccccccc}
\toprule
Method & UniFace & E4S & FaceDancer & FSGAN & InSwap & SimSwap & Avg.\\
\midrule
RECCE~\cite{recce} & 84.2 & 65.2 & 78.3 & 88.4 & 79.5 & 73.0 & 78.1\\
SBI~\cite{sbi} & 64.4 & 69.0 & 44.7 & 87.9 & 63.3 & 56.8 & 64.4\\
IID~\cite{iid} & 79.5 & 71.0 & 79.0 & 86.4 & 74.4 & 64.0 & 75.7\\
UCF~\cite{ucf} & 78.7 & 69.2 & 80.0 & 88.1 & 76.8 & 64.9 & 77.5\\
LSDA~\cite{lsda} & 85.4 & 68.4 & 75.9 & 83.2 & 81.0 & 72.7 & 77.8\\
CDFA~\cite{LinECCV2024CDFA} & 76.5 & 67.4 & 75.4 & 84.8 & 72.0 & 76.1 & 75.9\\
FIA-USA~\cite{ma2025from} & \textbf{91.8} & 87.5 & 83.0 & 86.3 & 87.4 & \textbf{91.0} & 87.8\\
ForAda~\cite{cui2025forensics} & 91.2 & \underline{92.9} & \underline{90.4} & \underline{94.7} & \textbf{90.0} & 85.6 & \underline{90.8}\\
\midrule
EISL (ours) & \underline{91.7} & \textbf{96.3} & \textbf{92.7} & \textbf{95.3} & \underline{87.8} & \underline{88.5} & \textbf{92.1}\\
\bottomrule
\end{tabular}
\end{table*}


\begin{table}[!tb]
\color{black}
\caption{Frame-level AUC (\%) on DiFF. All models are trained on FF++ and transferred without adaptation.}
\label{tab:diff}
\centering
\begin{tabular}{lcccc}
\toprule
Method & T2I & I2I & FS & FE\\
\midrule
Xception~\cite{ff++} & 62.43 & 56.83 & 85.97 & 58.64\\
F$^3$-Net~\cite{f3net} & 66.87 & 67.64 & 81.01 & 60.60\\
EfficientNet~\cite{efficientnet} & 74.12 & 57.27 & 82.11 & 57.20\\
DIRE~\cite{wang2023dire} & 44.22 & 64.64 & 84.98 & 57.72\\
SBI~\cite{sbi} & 80.20 & 80.40 & 85.08 & 68.79\\
FIA-USA~\cite{ma2025from} & \underline{86.05} & \underline{84.95} & \textbf{89.42} & \underline{72.73}\\
EISL (ours) & \textbf{89.86} & \textbf{86.21} & \underline{86.48} & \textbf{72.94}\\
\bottomrule
\end{tabular}
\end{table}


\noindent\textbf{DiFF Comparison}. \rev{Following~\cite{ma2025from}, Table~\ref{tab:diff} further evaluates the generalization ability from video-based manipulations in FF++ to whole-face synthesis and editing scenarios in DiFF~\cite{cheng2024diff}. Despite the substantial distribution gap between the training and testing forgery types, EISL achieves the best AUC on T2I (89.86), I2I (86.21), and facial editing (72.94), while remaining competitive on face swapping (86.48). In particular, the consistent improvements on both text-to-image and image-to-image synthesis indicate that the learned representation is not restricted to artifacts associated with the manipulation pipelines observed during training. These results further demonstrate the transferability of EISL to unseen generative paradigms and support its ability to capture more generalizable forgery-relevant representations.}

\subsection{Component Ablation}
\label{sec:ablation}
\begin{table*}[!t]
\caption{Component ablation. Metrics are frame-level AUC/AP/EER on CDF-v1, CDF-v2, and DFDCP. Ref., Inv., and Orth. denote $\mathcal{L}_{\mathrm{ref}}$, EIM with $\mathcal{L}_{\mathrm{inv}}$, and the low-rank projector with $\mathcal{L}_{\mathrm{orth}}$, respectively.}
\label{tab:ablation}
\centering
\setlength{\tabcolsep}{8pt}
\begin{tabular}{ccc|ccc|ccc|ccc|ccc}
\toprule
\multirow{2}{*}{Ref.} & \multirow{2}{*}{Inv.} & \multirow{2}{*}{Orth.} &
\multicolumn{3}{c|}{CDF-v1} & \multicolumn{3}{c|}{CDF-v2} & \multicolumn{3}{c|}{DFDCP} & \multicolumn{3}{c}{Average}\\
\cmidrule(lr){4-6}\cmidrule(lr){7-9}\cmidrule(lr){10-12}\cmidrule(lr){13-15}
&&&AUC&AP&EER&AUC&AP&EER&AUC&AP&EER&AUC&AP&EER\\
\midrule
$\times$&$\times$&$\times$&0.898&0.940&0.177&0.881&0.935&0.201&0.836&0.868&0.249&0.872&0.914&0.209\\
$\checkmark$&$\times$&$\times$&0.903&0.943&0.183&0.883&0.935&0.200&0.846&0.912&0.240&0.877&0.930&0.207\\
$\times$&$\checkmark$&$\times$&0.723&0.788&0.342&0.717&0.809&0.340&0.629&0.755&0.418&0.689&0.784&0.366\\
$\times$&$\times$&$\checkmark$&0.910&0.948&0.172&0.903&0.947&0.179&0.861&0.927&0.230&0.891&0.940&0.194\\
$\times$&$\checkmark$&$\checkmark$&0.933&0.961&0.145&0.906&0.948&0.178&0.870&0.930&0.217&0.903&0.943&0.180\\
$\checkmark$&$\times$&$\checkmark$&\textbf{0.934}&\textbf{0.962}&\textbf{0.145}&0.888&0.940&0.197&0.870&0.931&0.216&0.897&0.944&0.186\\
$\checkmark$&$\checkmark$&$\times$&0.907&0.945&0.178&0.891&0.940&0.189&0.866&0.927&0.224&0.888&0.937&0.197\\
$\checkmark$&$\checkmark$&$\checkmark$&0.920&0.953&0.162&\textbf{0.911}&\textbf{0.952}&\textbf{0.172}&\textbf{0.894}&\textbf{0.942}&\textbf{0.193}&\textbf{0.908}&\textbf{0.949}&\textbf{0.176}\\
\bottomrule
\end{tabular}
\end{table*}

\smallskip
\noindent\textbf{Effect of Components.}
We investigate the contributions of RefCLIP ($\mathcal{L}_{\text{ref}}$), EIM ($\mathcal{L}_{\text{inv}}$), and Subspace Layer ($\mathcal{L}_{\text{orth}}$) on CDF-v1, CDF-v2, and DFDCP, summarized in Table~\ref{tab:ablation}.
The baseline model without these components achieves an average AUC of 0.872, AP of 0.914, and EER of 0.209.
Adding RefCLIP alone slightly but consistently improves performance to an average AUC of 0.877, AP of 0.930, and EER of 0.207, which shows that alignment with the reference CLIP space provides more discriminative supervision for deepfake detection.
Using only EIM reduces the performance to an average AUC of 0.689, AP of 0.784, and EER of 0.366, which indicates that enforcing environment invariance in isolation is not sufficient, as environment perturbations alone are difficult to directly align with invariant information in the LoRA branch.
Introducing only the Subspace Layer increases the average AUC to 0.891, raises AP to 0.940, and lowers EER to 0.194, which confirms that an explicitly constrained subspace effectively disentangles low-rank  cues from the overall representation and enhances cross-dataset generalization.
When EIM is combined with the Subspace Layer, the average AUC and AP further rise to 0.903 and 0.943, and EER decreases to 0.180, which demonstrates that environment-invariant learning becomes beneficial once nuisance factors are restricted within a well-structured representation space.
The combination of RefCLIP and the Subspace Layer also surpasses the baseline and remains competitive, while the setting without the Subspace Layer shows inferior results compared with those that include it, highlighting the central role of the subspace constraint.
Using all three components achieves the best performance, showing the three losses are complementary and jointly contribute to robust forgery detection.

\begin{table}[!tb]
\color{black}
\caption{Projection structure. Frame-level AUC obtained by keeping all objectives fixed and replacing only the mapping applied to $F$.}
\label{tab:structure}
\centering
\begin{tabular}{lccc}
\toprule
Mapping & CDF-v1 & CDF-v2 & DFDCP\\
\midrule
No projection ($P=I_D$) & 0.907 & 0.891 & 0.866\\
Full-rank linear ($D\times D$) & 0.899 & 0.891 & 0.885\\
Two linear layers & 0.891 & 0.894 & 0.868\\
EISL projector ($QQ^\top$, $r=64$) & \textbf{0.920} & \textbf{0.911} & \textbf{0.894}\\
\bottomrule
\end{tabular}
\end{table}

\noindent\textbf{\rev{Projection Structure.}}
\label{sec:structure}
\rev{Table~\ref{tab:structure} compares different feature mappings while keeping the backbone, training protocol, and all optimization objectives unchanged. The ``No projection'' variant directly applies the invariance objective in the original feature space, while the full-rank linear layer and two-layer mapping test whether the improvement can be attributed simply to introducing additional trainable transformations or greater mapping capacity. As shown in Table~\ref{tab:structure}, neither alternative consistently matches the proposed low-rank $QQ^\top$ projector, which achieves the best AUC on all three datasets. This indicates that the gain of EISL does not arise merely from adding trainable parameters or enforcing $\mathcal{L}_{\mathrm{inv}}$ on the original representation, but from the constrained projection geometry that restricts features to a compact subspace and suppresses environment-sensitive directions.}

\smallskip
\noindent\textbf{\rev{Projector Rank and Backbone.}}
\begin{table*}[!t]
\color{black}
\caption{Effect of the projector rank $r$ and of the CLIP backbone (frame-level AUC/AP/EER). Each block varies one factor while holding the other at its default value, so the default configuration ($r=64$, ViT-L/14) is the same run in both blocks.}
\label{tab:rank_backbone}
\centering
\setlength{\tabcolsep}{8pt}
\begin{tabular}{l|ccc|ccc|ccc|ccc}
\toprule
\multirow{2}{*}{Variant} & \multicolumn{3}{c|}{CDF-v1} & \multicolumn{3}{c|}{CDF-v2} & \multicolumn{3}{c|}{DFDCP} & \multicolumn{3}{c}{Average}\\
\cmidrule(lr){2-4}\cmidrule(lr){5-7}\cmidrule(lr){8-10}\cmidrule(lr){11-13}
&AUC&AP&EER&AUC&AP&EER&AUC&AP&EER&AUC&AP&EER\\
\midrule
\multicolumn{13}{l}{\textit{Projector rank $r$ (backbone fixed to ViT-L/14)}}\\
\quad $r=32$ & 0.921 & 0.951 & 0.155 & 0.884 & 0.933 & 0.197 & 0.886 & 0.931 & 0.203 & 0.897 & 0.938 & 0.185\\
\quad $r=64$ (default) & 0.920 & 0.953 & 0.162 & \textbf{0.911} & \textbf{0.952} & \textbf{0.172} & \textbf{0.894} & \textbf{0.942} & \textbf{0.193} & \textbf{0.908} & \textbf{0.949} & \textbf{0.176}\\
\quad $r=128$ & 0.908 & 0.945 & 0.167 & 0.898 & 0.941 & 0.182 & 0.872 & 0.929 & 0.219 & 0.892 & 0.938 & 0.189\\
\midrule
\multicolumn{13}{l}{\textit{CLIP backbone (rank fixed to $r=64$)}}\\
\quad ViT-B/32 & 0.807 & 0.877 & 0.265 & 0.816 & 0.884 & 0.263 & 0.769 & 0.860 & 0.304 & 0.797 & 0.874 & 0.277\\
\quad ViT-B/16 & 0.860 & 0.919 & 0.216 & 0.815 & 0.894 & 0.271 & 0.862 & 0.919 & 0.233 & 0.845 & 0.910 & 0.240\\
\quad ViT-L/14 (default) & \textbf{0.920} & \textbf{0.953} & \textbf{0.162} & \textbf{0.911} & \textbf{0.952} & \textbf{0.172} & \textbf{0.894} & \textbf{0.942} & \textbf{0.193} & \textbf{0.908} & \textbf{0.949} & \textbf{0.176}\\
\bottomrule
\end{tabular}
\end{table*}
\rev{Table~\ref{tab:rank_backbone} investigates the effects of the projector rank and the CLIP visual backbone. With ViT-L/14 fixed, the performance exhibits a non-monotonic trend as the rank increases. Specifically, $r=64$ achieves the best overall results, with average AUC, AP, and EER values of 0.908, 0.949, and 0.176, respectively. Compared with $r=32$, it improves the average AUC and AP by 1.1 percentage points and reduces the EER by 0.9 percentage points. The gains mainly come from CDF-v2 and DFDCP: on CDF-v2, $r=64$ improves AUC/AP by 2.7/1.9 points and reduces EER by 2.5 points, while on DFDCP it improves AUC/AP by 0.8/1.1 points and reduces EER by 1.0 point. Although $r=32$ is marginally better on CDF-v1 in terms of AUC and EER, $r=64$ provides more stable performance across the three test domains. Increasing the rank further to 128 degrades the average AUC from 0.908 to 0.892 and increases the EER from 0.176 to 0.189. The degradation is especially evident on DFDCP, where the AUC drops from 0.894 to 0.872 and the EER rises from 0.193 to 0.219. These results indicate that increasing the projector capacity does not necessarily improve cross-dataset detection. A small rank may overly constrain feature adaptation, whereas a large rank may retain more dataset-specific variations. Therefore, $r=64$ offers a better balance between compactness and transferability and is adopted as the default setting. With $r=64$ fixed, ViT-L/14 consistently outperforms both ViT-B backbones on all datasets and metrics. Compared with ViT-B/16, it improves the average AUC and AP by 6.3 and 3.9 percentage points and reduces the EER by 6.4 points. Compared with ViT-B/32, the corresponding improvements are 11.1, 7.5, and 10.1 points. The gains are particularly pronounced on CDF-v2 and DFDCP; for example, ViT-L/14 improves the CDF-v2 AUC over ViT-B/16 from 0.815 to 0.911 and the DFDCP AUC over ViT-B/32 from 0.769 to 0.894. This suggests that the stronger encoder provides more transferable representations for detecting manipulation traces under distribution shifts. Nevertheless, the proposed projector remains applicable to both ViT-B variants without architectural modification, showing that EISL is not tied to a specific CLIP encoder. Finally, the backbone has a substantially larger impact than the projector rank: replacing ViT-B/32 with ViT-L/14 improves the average AUC by 11.1 points, whereas changing the rank from 32 to 64 yields a 1.1-point gain. Thus, the backbone primarily determines representation quality, while the projector rank adjusts the trade-off between information preservation and cross-dataset generalization.}

\smallskip
\noindent\textbf{\rev{Intervention Mask Design.}}
\begin{table}[!t]
\color{black}
\caption{Effect of mask target and shape (frame-level AUC).}
\label{tab:mask}
\centering
\setlength{\tabcolsep}{4pt}
\begin{tabular}{lccc}
\toprule
Mask & CDF-v1 & CDF-v2 & DFDCP\\
\midrule
Background only, fixed & 0.905 & 0.899 & 0.886\\
Foreground/background, fixed & 0.910 & 0.907 & \textbf{0.895}\\
Foreground/background, variable & \textbf{0.920} & \textbf{0.911} & 0.894\\
\bottomrule
\end{tabular}
\end{table}
\rev{Table~\ref{tab:mask} evaluates two design choices of the intervention mask in Eq.~\eqref{eq:mask}, namely the intervention target and the mask geometry. Applying a fixed mask only to the background yields the lowest average AUC of 0.897. Extending the intervention to alternate between the sampled region and its complement improves the AUC on CDF-v1, CDF-v2, and DFDCP by 0.5, 0.8, and 0.9 percentage points, respectively, indicating that interventions covering both facial and contextual regions provide more informative variation than background-only perturbations. Further randomizing the mask shape and spatial extent increases the average AUC from 0.904 to 0.908 and achieves the best results on CDF-v1 and CDF-v2. On DFDCP, the fixed and variable designs perform comparably, with the fixed mask being marginally better by 0.1 percentage points. Overall, these results suggest that varying the intervention geometry improves cross-dataset robustness, potentially by reducing the model's dependence on fixed mask boundaries and spatial locations. We therefore adopt the variable foreground/background mask as the default design.}

\smallskip
\noindent\textbf{\rev{Loss-Weight Sensitivity.}}
\begin{table}[!t]
\color{black}
\caption{Sensitivity to the loss weights in Eq.~\eqref{eq:total}. One coefficient is varied at a time, and the other two remain at their default values.}
\label{tab:loss_weight}
\centering
\footnotesize
\setlength{\tabcolsep}{5pt}
\begin{tabular}{ccrrrr}
\toprule
Weight & Value & CDF-v1 & CDF-v2 & DFDCP & Avg.\\
\midrule
$\lambda_{\mathrm{ref}}$  & $1$    & 0.828 & 0.798 & 0.804 & 0.810\\
                             & $0.5$  & 0.874 & 0.882 & 0.871 & 0.876\\
                             & $0.2$  & 0.916 & 0.886 & \textbf{0.896} & 0.899\\
                             & $0.1$  & \textbf{0.920} & \textbf{0.911} & 0.894 & \textbf{0.908}\\
\midrule
$\lambda_{\mathrm{inv}}$  & $1$    & 0.919 & 0.907 & 0.893 & 0.906\\
                             & $0.5$  & 0.910 & 0.896 & 0.886 & 0.897\\
                             & $0.2$  & \textbf{0.920} & \textbf{0.911} & 0.894 & \textbf{0.908}\\
                             & $0.1$  & 0.916 & 0.889 & \textbf{0.899} & 0.901\\
\midrule
$\lambda_{\mathrm{orth}}$ & $1$    & 0.914 & 0.897 & 0.891 & 0.901\\
                             & $0.1$  & 0.913 & 0.894 & 0.890 & 0.899\\
                             & $0.01$ & \textbf{0.920} & \textbf{0.911} & \textbf{0.894} & \textbf{0.908}\\
\bottomrule
\end{tabular}
\end{table}
\rev{Table~\ref{tab:loss_weight} studies the sensitivity of EISL to the three objective coefficients by varying one coefficient at a time while keeping the other two fixed at their default values. Across the three independent sweeps, the best average AUC is observed at $\lambda_{\mathrm{ref}}=0.1$, $\lambda_{\mathrm{inv}}=0.2$, and $\lambda_{\mathrm{orth}}=0.01$. For $\lambda_{\mathrm{ref}}$, the performance peaks at 0.1 and decreases when a larger weight is used, suggesting that excessive regularization toward the reference CLIP representation may limit task-specific adaptation. Similarly, $\lambda_{\mathrm{inv}}=0.2$ achieves the most favorable performance, indicating that a moderate invariance constraint can suppress environment-specific variations while preserving forgery-relevant information. For the orthogonality term, $\lambda_{\mathrm{orth}}=0.01$ provides the best result. At this setting, $\mathcal{L}_{\mathrm{orth}}$ decreases to approximately $10^{-4}$, indicating that the desired orthogonality has already been effectively enforced. Assigning a larger weight brings no further improvement and instead degrades cross-dataset performance, possibly because an overly strong constraint interferes with the early-stage optimization of the discriminative representation. Based on these results, we adopt $\lambda_{\mathrm{ref}}=0.1$, $\lambda_{\mathrm{inv}}=0.2$, and $\lambda_{\mathrm{orth}}=0.01$ in all other experiments.}

\subsection{\rev{Environmental Sensitivity and Mechanism Analysis}}
\label{sec:mechanism}
\rev{The preceding results establish the generalization gains of EISL. This subsection connects these gains to the mechanism in Sec.~\ref{sec:subspace} through four complementary analyses: environmental sensitivity under label-preserving perturbations, the routing of intervention-induced variation across the projected and residual components, semantic retention in the invariant subspace, and controlled attribution against the CLIP-LoRA baseline.}

\noindent\textbf{\rev{Environmental Sensitivity of the Adapted Detector}}
\label{sec:env_sensitivity}
\begin{table}[!t]
\color{black}
\caption{Environmental sensitivity of the CLIP-LoRA baseline. A single label-preserving perturbation is applied to the test images; the forgery source, identity, and label are unchanged. Frame-level AUC.}
\label{tab:envperturb}
\centering
\begin{tabular}{lcccc}
\toprule
Perturbation & CDF-v1 & CDF-v2 & DFDCP & Avg.\\
\midrule
None & 0.898 & 0.881 & 0.836 & 0.872\\
Gaussian blur & 0.857 & 0.848 & 0.803 & 0.836\\
Grayscale & 0.879 & 0.853 & 0.771 & 0.834\\
Autocontrast & 0.824 & 0.755 & 0.729 & 0.769\\
\bottomrule
\end{tabular}
\end{table}
\rev{Table~\ref{tab:envperturb} isolates environmental sensitivity on the CLIP-LoRA baseline, which is trained without interventions. With the forgery source, identity, and label fixed, Gaussian blur, grayscale conversion, and autocontrast reduce the average AUC from 0.872 to 0.836, 0.834, and 0.769, corresponding to drops of 3.6, 3.8, and 10.3 points, respectively. The largest degradation is nearly three times the 3.6-point clean-data gain of the complete method over the same baseline in Table~\ref{tab:ablation}.}

\rev{These results demonstrate that environmental variation alone produces substantial forensic errors and constitutes an independent source of cross-domain degradation beyond generator shift. Because its effect can be directly measured under label-preserving transformations, environmental variation can be explicitly modeled and suppressed through controlled interventions. Together with generator-specific artifacts and identity bias discussed in Sec.~\ref{sec:scope}, environmental variation forms a major source of cross-domain failure addressed by EISL.}

\noindent\textbf{\rev{Intervention Response of the Projected and Residual Subspaces}}
\label{sec:intervention_response}
\begin{table}[!t]
\color{black}
\caption{Intervention response on CDF-v2. Shift is the mean intervention-induced feature displacement, normalized by the distance between the original real and fake feature means.}
\label{tab:intervention}
\centering
\begin{tabular}{lccc}
\toprule
Feature & Inter.\ AUC & Drop & Shift\\
\midrule
CLIP-LoRA & 0.772 & 10.9 pt & 1.99$\times$\\
EISL, projected $\widetilde F$ & \textbf{0.866} & \textbf{4.5 pt} & \textbf{0.66}$\times$\\
EISL, residual $(I_D-P)F$ & -- & -- & 2.63$\times$\\
\bottomrule
\end{tabular}
\end{table}
\begin{figure*}[!t]
    \centering
    \includegraphics[width=\textwidth]{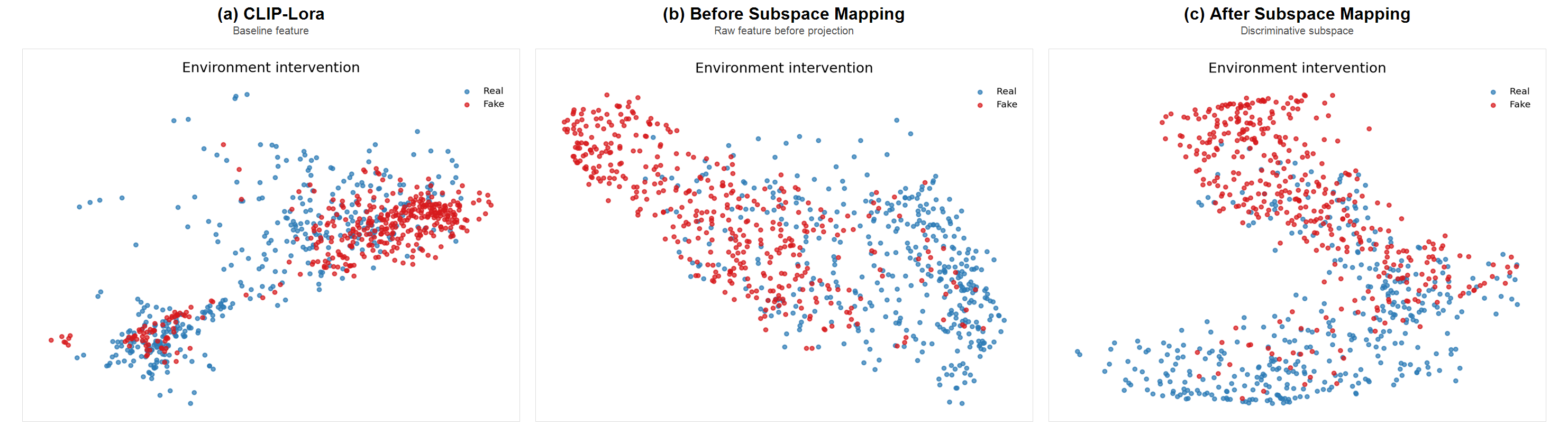}
    \caption{\rev{t-SNE visualization on CDF-v2 after environmental intervention. The CLIP-LoRA baseline is strongly entangled. EISL features before projection still contain intervention-sensitive variation, whereas the projected subspace provides clearer real/fake separation.}}
    \label{fig:intervention_tsne}
\end{figure*}
\begin{figure*}[!t]
    \centering
    \includegraphics[width=\textwidth]{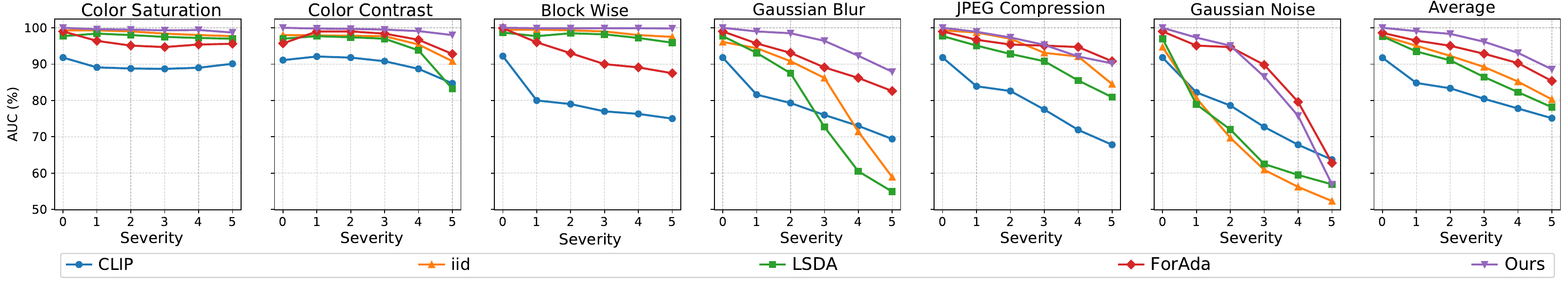}
    \caption{Video-level AUC under five severity levels of six perturbations. EISL is compared with CLIP, IID~\cite{iid}, LSDA~\cite{lsda}, and ForAda~\cite{cui2025forensics}.}
    \label{fig:robustness}
\end{figure*}
\rev{Proposition~3 states that intervention-induced variation should be pushed out of $\mathcal{S}$ and carried by the residual component. Table~\ref{tab:intervention} verifies this mechanism on CDF-v2 under a regional environmental intervention. Shift denotes the mean intervention-induced feature displacement normalized by the distance between the original real and fake feature means, thereby measuring the magnitude of environmental variation relative to the original inter-class structure. A value above $1\times$ indicates that the intervention moves a sample farther than the separation between the two class centers and therefore substantially disrupts the original real/fake discriminative structure. For CLIP-LoRA, the intervention reduces AUC from 0.881 to 0.772, a drop of 10.9 points, while the shift reaches $1.99\times$. Thus, environmental variation alone induces a feature displacement nearly twice the original class-center separation, substantially weakening the separability between real and fake samples. In the EISL projected subspace, AUC decreases from 0.911 to 0.866, corresponding to a drop of only 4.5 points, while the shift is reduced to $0.66\times$, approximately $67\%$ lower than that of CLIP-LoRA. The smaller feature displacement and AUC degradation jointly demonstrate that the projected representation is substantially more stable under environmental variation. Conversely, the residual shift increases to $2.63\times$, exhibiting the opposite response to that of the projected subspace. These complementary responses reveal the underlying mechanism of EISL: intervention-induced variation is suppressed in the retained subspace and routed into the complementary residual component, rather than being indiscriminately erased during projection. The simultaneous reduction in projected shift and increase in residual shift directly accord with the orthogonal decomposition described in Proposition~3 and Eq.~\eqref{eq:orth_decomp}, demonstrating that EISL performs a structured separation of environment-sensitive variation from task-relevant information rather than generic feature compression. Moreover, the projected features are more stable than the CLIP representation to which they are semantically aligned. Figure~\ref{fig:intervention_tsne} provides the corresponding qualitative evidence: EISL features before projection retain pronounced intervention-sensitive spread, whereas the projected features suppress this spread and yield a markedly clearer separation between real and fake samples, consistent with the quantitative results.}



\subsection{Robustness to Real-World Corruptions}
\label{sec:corruption}

\rev{Because EIM is used during training, we test whether its benefit transfers to degradations outside the intervention pool.} Following DeeperForensics~\cite{DeeperForensics}, we evaluate color saturation, contrast, block-wise artifacts, Gaussian blur, JPEG compression, and Gaussian noise at five severity levels. \rev{Block-wise artifacts, JPEG compression, and Gaussian noise are structurally different from the photometric and smoothing operations in $\mathcal{A}$.} Fig.~\ref{fig:robustness} shows that EISL performs best under block-wise artifacts, blur, compression, and noise, and remains competitive under saturation and contrast changes. Averaged over all distortion types and levels, it ranks first. \rev{The advantage on corruption families outside the intervention pool suggests that the learned subspace suppresses a broader class of appearance-driven directions rather than memorizing the operations used to construct the training pairs.}

\section{\rev{Discussion and Limitations}}
\label{sec:scope}

\rev{Our study focuses on environmental variations that preserve the underlying forensic target while altering the visual conditions under which it is observed. Following Sec.~\ref{sec:env_def}, we define environmental factors as label-preserving, non-forgery variations arising from acquisition, post-processing, or visual context. EIM provides a controllable instantiation of this family through photometric, degradation, and context/style changes, following the intervention paradigm used for invariant analysis of vision--language models~\cite{Song_2024_CLIPICM}. These variations reflect appearance changes commonly introduced when media are captured, processed, or redistributed, without modifying the depicted identity, forgery source, or real/fake label.}

\rev{The experimental results provide consistent evidence that such environmental variations constitute a meaningful source of cross-domain degradation. As shown in Sec.~\ref{sec:env_sensitivity}, label-preserving perturbations alone can reduce the AUC of the CLIP-based detector by up to 10.3 points even when the forgery source remains unchanged. Sec.~\ref{sec:intervention_response} further shows that, after subspace projection, intervention-induced variation is substantially reduced in the retained representation and concentrated in the complementary residual component. Moreover, the robustness improvements on corruption families outside the intervention pool in Sec.~\ref{sec:corruption} suggest that EISL does not simply memorize the predefined intervention operations, but suppresses a broader set of appearance-sensitive feature directions. Together, these observations support the hypothesis that a non-trivial portion of CLIP detector sensitivity can be attributed to suppressible environment-dependent variations.}

\rev{Nevertheless, environmental variation represents only one source of distribution shift in open-world deepfake detection. The current intervention family does not explicitly model every possible acquisition condition or generator-specific artifact, since changing the underlying manipulation mechanism would violate the label-preserving intervention criterion. Other factors, including identity bias, dataset curation effects, and benchmark-specific processing pipelines, may also contribute to the remaining generalization gap. Therefore, the analysis in Sec.~\ref{sec:mechanism} should be interpreted as identifying and mitigating a measurable subset of nuisance variations, rather than providing a complete decomposition of all domain-dependent factors.}

\rev{The current subspace formulation also adopts a deliberately simple design, using a single fixed-rank linear projector shared across samples and environments. This formulation introduces few additional parameters and provides an explicit geometric interpretation through the projected and residual components, as analyzed in Sec.~\ref{sec:subspace}. However, a global linear subspace may be insufficient to capture nonlinear interactions among identity, content, forgery evidence, and environmental conditions, while a fixed rank cannot account for sample-dependent variations in nuisance complexity. Richer semantic or generative interventions, adaptive intervention policies, nonlinear invariant subspaces, and sample-dependent or adaptive-rank projections therefore represent promising directions for future work.}
\section{Conclusion}
\label{sec:conclusion}
\rev{We presented EISL, which uses label-preserving environmental interventions to identify unstable CLIP feature directions and a low-rank orthogonal projection to exclude them from the discriminative subspace. The geometric and rank properties of the projector are linked to paired consistency in the method analysis and tested through reduced projected shift, increased residual shift, retained identity semantics, and weaker alternatives based on identity, full-rank, or unconstrained mappings. Cross-dataset, cross-generator, whole-face synthesis, and corruption experiments show consistent gains, with the largest margins under stronger train--test shifts. As bounded in Sec.~\ref{sec:scope}, the current evidence covers the implemented intervention family and a single fixed-rank linear subspace; richer interventions and nonlinear or adaptive-rank projections are promising extensions.}

\FloatBarrier
\def\IEEEbibitemsep{-2.85pt}
\bibliographystyle{IEEEtran}
\bibliography{references}


\end{document}